\documentclass[conference]{IEEEtran}
\IEEEoverridecommandlockouts
\usepackage{cite}
\usepackage{amsmath,amssymb,amsfonts}
\usepackage{graphicx}
\usepackage{subcaption}
\usepackage{tabularx}
\usepackage{multirow}
\usepackage{textcomp}
\usepackage{xcolor}
\usepackage{float}
\usepackage{verbatim}
\usepackage{url}

\usepackage[english]{babel}
\usepackage[utf8]{inputenc}
\usepackage{algorithm}
\usepackage[noend]{algpseudocode}

\def\BibTeX{{\rm B\kern-.05em{\sc i\kern-.025em b}\kern-.08em
    T\kern-.1667em\lower.7ex\hbox{E}\kern-.125emX}}
\begin{document}

\title{Multi-view autoencoders for Fake News Detection \thanks{This work was partially funded by Brazilian agencies: FACEPE and CNPq.}}

\author{\IEEEauthorblockN{Ingryd V. S. T. Pereira}
\IEEEauthorblockA{\textit{Centro de Informática} \\
\textit{Universidade Federal de Pernambuco}\\
Recife-PE, Brazil \\
ivstp@cin.ufpe.br}
\and
\IEEEauthorblockN{George D. C. Cavalcanti}
\IEEEauthorblockA{\textit{Centro de Informática} \\
\textit{Universidade Federal de Pernambuco}\\
Recife-PE, Brazil \\
gdcc@cin.ufpe.br}
\and
\IEEEauthorblockN{Rafael M. O. Cruz}
\IEEEauthorblockA{\textit{École de technologie supérieure} \\
\textit{Université du Québec}\\
Montreal, Canada \\
rafael.menelau-cruz@etsmtl.ca}
}


\maketitle

\begin{abstract}
Given the volume and speed at which fake news spreads across social media, automatic fake news detection has become a highly important task. However, this task presents several challenges, including extracting textual features that contain relevant information about fake news. Research about fake news detection shows that no single feature extraction technique consistently outperforms the others across all scenarios. Nevertheless, different feature extraction techniques can provide complementary information about the textual data and enable a more comprehensive representation of the content. This paper proposes using multi-view autoencoders to generate a joint feature representation for fake news detection by integrating several feature extraction techniques commonly used in the literature. Experiments on fake news datasets show a significant improvement in classification performance compared to individual views (feature representations). We also observed that selecting a subset of the views instead of composing a latent space with all the views can be advantageous in terms of accuracy and computational effort. For further details, including source codes, figures, and datasets, please refer to the project’s repository: \url{https://github.com/ingrydpereira/multiview-fake-news}.
\end{abstract}


\section{Introduction}
Fake news detection has always been a challenging task, even more so when performed manually by humans. After the popularization of social media, the amount of news generated and the speed at which it reaches end users increased significantly. Hence, the importance of this task be performed automatically \cite{guo2022survey}.  Although machine learning techniques have been widely applied to this problem, their effectiveness heavily depends on the quality of the data representation. Many traditional models struggle because they rely on shallow features, such as word counts or basic linguistic patterns, which fail to capture the deeper semantics and contextual nuances present in fake news. 
Therefore, there is a need for more advanced models that can learn deeper, context-aware representations of textual data.

One of the most considerable difficulties in fake news detection in texts is extracting features.
There are several ways to represent texts, each with its particularities, and some ways may not contain all the relevant information of the text. This is a problem for all tasks involving texts, but in fake news detection, it becomes ever more difficult due to the limitation of news contents and the nature of the data, which is created to deceive the reader \cite{lazer2018science, pennycook2021psychology}.

The study conducted by Farhangian et al. \cite{farhangian2024fake} provides a comprehensive assessment of the current state of Fake News detection, utilizing well-established classifiers and feature extraction techniques from the literature. In addition to offering a comparative analysis of these methods, their findings emphasize not only the diversity of text feature extraction techniques but also the complementary nature of these representations. This complementary relationship is the most significant insight from the survey and serves as a key motivation for this paper, highlighting how combining various feature extraction methods can enhance the effectiveness of fake news detection.

Extensive research has been conducted on the topic of fake news, exploring diverse multi-view approaches. These include the use of different modalities, such as text and image \cite{ying2023bootstrapping}; views on different aspects of the news, such as the content itself, the source, and the user who posted the news \cite{ni2021mvan, bazmi2023multi}; perspectives on different facets of news interpretation, such as semantics, emotional tone, and style \cite{zhu2022memory}; and even approaches based on semantic relations of text-scenes and text-objects derived from graphs \cite{cui4411791multi}. However, no research has been found that employs different textual feature extraction methods as views in fake news detection models within a multi-view framework. Despite the complementary nature of these representations, as highlighted in~\cite{farhangian2024fake}, this gap in the literature motivates us to explore approaches for creating a multi-view system based solely on diverse text representations.

We propose a fake news detection method using a multi-view autoencoder, which integrates multiple feature views into a single joint representation to enhance feature representation. By leveraging the autoencoder's ability to learn robust features enriched with information from all views, this approach aims to provide a comprehensive understanding of the underlying structures and patterns in fake news content. We hypothesize that this enhanced representation improves detection performance compared to traditional single-view methods.

To validate this, we compare the proposed model with fake news detection techniques baselines. We conduct experiments with four datasets, seven multi-view autoencoder models, and seven classifiers to answer the following research questions: (1)~Can any multi-view autoencoder model create a new feature representation that contains enough information about fake news? (2)~Can a multi-view approach represent features better than individual views? (3)~Do all views contain relevant information or are just a few of these views enough?

In summary, our contributions are as follows:
\begin{itemize}
  \item A novel approach for fake news detection using multi-view autoencoders.
  \item An comparative analysis of the proposal with different multi-view autoencoder models.
  \item An evaluation of the impact of different sets of views.
\end{itemize}

\section{Multi-view autoencoders}

\begin{figure}[]
\centering
\begin{subfigure}[t]{0.4\textwidth}
  \centering
  \includegraphics[width=\linewidth]{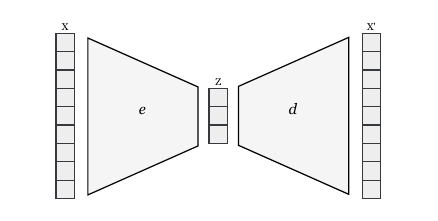}
  \caption{Autoencoder}
  \label{fig:AE}
\end{subfigure}
\begin{subfigure}[t]{0.4\textwidth}
  \centering
  \includegraphics[width=\linewidth]{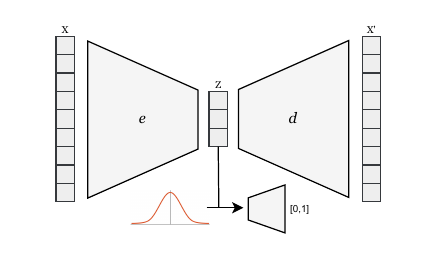}
  \caption{Adversarial Autoencoder}
  \label{fig:AAE}
\end{subfigure}
\begin{subfigure}[t]{0.4\textwidth}
  \centering
  \includegraphics[width=\linewidth]{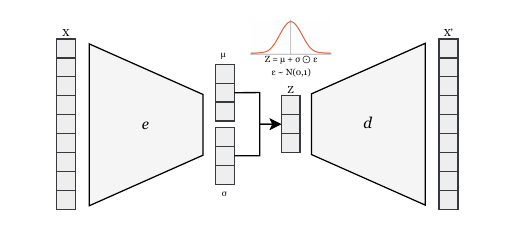}
  \caption{Variational Autoencoder}
  \label{fig:VAE}
\end{subfigure}
\caption{Single view autoencoder (adapted from Aguila et al.~\cite{aguila2023multi})}
\label{fig:all_clfs_by_multiae}
\end{figure}

Autoencoder is an unsupervised model in which the input and the output layers have the same number of attributes/dimensions. Its objective is to reconstruct the data while preserving key properties of the input, after reducing its dimensionality. Figure~\ref{fig:AE} contains a simple abstraction of this model. An autoencoder has both an encoder ($e$) and a decoder ($d$) in the structure. The encoder ($e$) aims to condense the input data ($X$) into a smaller representation, named as latent vector ($Z$). The decoder ($d$) takes as input the latent vector ($Z$), and reconstruct the original data ($X'$) \cite{deng2014deep}. 

\begin{figure}[]
  \centering
  \includegraphics[width=\linewidth]{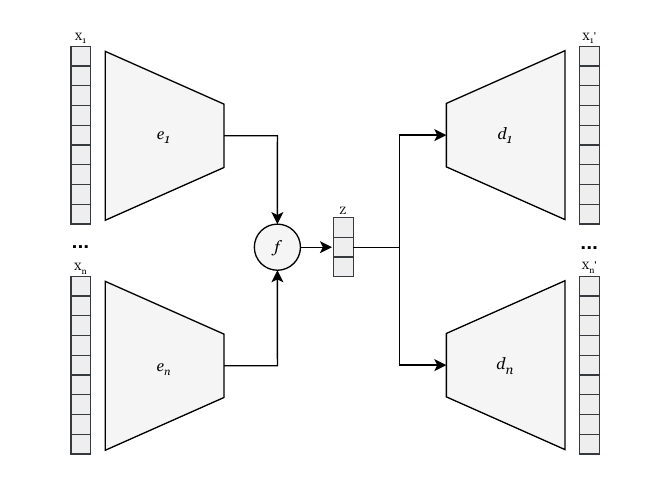}
  \caption{Multi-view Autoencoder}
  \label{fig:multiAE}
\end{figure}

The multi-view autoencoder employs this same logic, but take multiple views as input, reducing all information and then reconstructing the original data, i.e., all individual views. Figure \ref{fig:multiAE} presents an abstraction of this model. One approach to the multi-view autoencoder is that the latent layer condenses all views ($[X_1, \dots, X_n]$) into a single data vector ($Z$).
This unified representation contains information about all views, allowing them to be reconstructed individually from this same vector.

The multi-view autoencoder model has an encoder ($[e_1, \dots, e_n]$) and a decoder ($[d_1, \dots, d_n]$) for each view ($n$). Each decoder receives the unified representation of the latent space ($Z$) as input and then reconstructs the original view ($[X'_1, \dots, X'_n]$). There is also one encoder for each view, but the construction of the latent space depends on the type of multi-view autoencoder. Generally, there are two types: adversarial (Figure~\ref{fig:AAE}) and variational (Figure~\ref{fig:VAE}) autoencoders.

In Multi-view Adversarial Autoencoders, the latent vector ($Z$) is the mean of mapped outputs of individual encoders. It uses a discriminator, which is a classifier, to differentiate fake samples from samples generated by the encoder, and its output helps to calibrate the encoders. Some models of this type are jointAAE (Multi-view Adversarial Autoencoder with joint latent representation) and wAAE (Multi-view Adversarial Autoencoder with joint latent representation and Wasserstein loss) \cite{wang2019adversarial}.

In Multi-view Variational Autoencoders, the encoders produce a mean ($\mu$) and a variance ($\sigma$) vector, which parameterizes a multivariate Gaussian distribution to represent the latent variables ($Z$) of each view. The step that builds the joint representation of the latent vector ($Z$) is called sampling, and it is built based on the mean ($\mu$) and variance ($\sigma$) variables, but its construction has different approaches and depends on the model type. Some variational models are mVAE (Multimodal Variational Autoencoder) \cite{wu2018multimodal}, DVCCA (Deep Variational CCA) \cite{wang2016deep}, me\_mVAE (Multimodal Variational Autoencoder with separate ELBO terms for each view) \cite{wu2018multimodal}, MoPoEVAE (Mixture-of-Products-of-Experts VAE) \cite{sutter2021generalized}, and mvtCAE (Multi-View Total Correlation Auto-Encoder) \cite{hwang2021multi}.

\section{Proposed Methodology}

We propose using multi-view autoencoders to fuse existing views into a more robust representation, which serves as input to a classifier for fake news detection. 
This proposal is structured into three primary steps: the training of the multi-view autoencoder, the training of the classifier, and the prediction. The multi-view autoencoder aims to extract a data representation from a multi-view input set and later use this information as input for a classifier.
The multi-view autoencoder consists of a set of encoders ($\mathbf{e}$), a set of decoders ($\mathbf{d}$) and a joint function ($f$), where $\mathbf{e} = [e_1, e_2,\dots, e_n]$ and $\mathbf{d} = [d_1, d_2,\dots, d_n]$, considering that $n$ is the number of views. 

Algorithm~\ref{alg:train-multiae} shows the multi-view autoencoder training process. This process needs to input a data set ($X$) and a joint function ($f$) specific to the multi-view autoencoders model. For each of the $n$ views, features are extracted from the input data $X$ using a view-specific feature extraction function, resulting in $X_v$ (line 4). These features are then encoded by the corresponding encoder $e_v$ to produce latent representations $Z_v$ (line 5). The latent representations from all views are subsequently fused using a joint function $f$ to obtain a combined latent space representation $Z$ (line 6). Next, for each view, the combined latent representation $Z$ is decoded by the corresponding decoder $d_v$ to reconstruct the original input features $X'_v$ (line 8). The reconstruction loss $\mathcal{L}$ is computed by comparing the original input features $\{X_1, \dots, X_n\}$ with their reconstructions $\{X'_1,  \dots, X'_n\}$ (line 9), and the calculation is defined by multi-view autoencoder type. This loss is used to update the parameters of the encoder $e_v$ and decoder $d_v$ through backpropagation (lines 11-12). The process continues iteratively until a stopping condition is met, at which point the function returns the trained set of encoders $\mathbf{e}$ (line 14). Although the autoencoder model comprises sets of encoders and decoders, we only return the encoders because our proposed solution utilizes only the dimensionality reduction stage of the autoencoder.

The classifier training process (Algorithm~\ref{alg:train-clf}) starts by extracting features from the input data $X$ for each view $v$ using the view-specific feature extraction function, resulting in the feature set $X_v$ (line 3). These features are then passed through the corresponding trained encoder $e_v$ to generate the latent representations $Z_v$ (line 4). All views' latent representations are combined using the joint function $f$, producing a unified latent space representation $Z$ (line 5). This joint latent representation $Z$ is then used as input to train the classifier model $\lambda$ (line 6), which is returned in line 7.

The prediction process (Algorithm~\ref{alg:prediction}) receives as input a query instance $x_q$, the set of encoders ($\mathbf{e}$), and the joint function $f$. The first step is to represent the query sample $x_q$ in $n$ different views using the feature extraction functions ($\text{feature-extraction}_v$) (line 3). These extracted features ($x_v$) are then encoded by the pre-trained encoders $e_v$ to produce the latent representations $z_v$ (line 4). The individual latent representations from all views are combined using the joint function $f$ to create a unified latent vector $z$ (line 5). This unified representation is the input to the trained classifier $\lambda$ to generate the predicted output $\hat{y}$ (line 6). The function concludes by returning the prediction $\hat{y}$ (line 7), which represents the classifier's inference based on the given multi-view input data.

\begin{algorithm}
\caption{Multi-view autoencoders training}\label{alg:train-multiae}
\begin{algorithmic}[1]
\Function{TrainMultiViewAutoencoder}{$X$, $f$}
    \While{!stop-condition}
        \For{view $v$ = 1 to $n$}
            \State $X_v = \text{feature-extraction}_v(X)$
            \State $Z_v = e_v(X_v)$ \Comment{$e_v$ is an encoder}
        \EndFor
        \State $Z = f(Z_1, Z_2, \dots, Z_n)$ \Comment{$f$ is the joint function}
        \For{view $v$ = 1 to $n$}
            \State $X'_v = d_v(Z)$ \Comment{$d_v$ is a decoder}
        \EndFor
        \State $\mathcal{L} = \text{calculate-loss}(\{X_1, \dots, X_n\}, \{X'_1,  \dots,X'_n\})$
        \For{view $v$ = 1 to $n$}
            \State $e_v = \text{train-encoder}(\mathcal{L}, e_v)$
            \State $d_v = \text{train-decoder}(\mathcal{L}, d_v)$
        \EndFor
       \State $\mathbf{e} = [e_1, e_2, \dots, e_n]$
    \EndWhile
    \State \textbf{Return} $\mathbf{e}$
\EndFunction
\end{algorithmic}
\end{algorithm}

\begin{algorithm}
\caption{Classifier training}\label{alg:train-clf}
\begin{algorithmic}[1]
\Function{TrainClassifier}{$X$, $\mathbf{e}$, $f$}
        \For{view $v$ = 1 to $n$}
            \State $X_v = \text{feature-extraction}_v(X)$
            \State $Z_v = e_v(X_v)$ 
        \EndFor
        \State $Z = f(Z_1, Z_2, \dots, Z_n)$ 
        \State $\lambda = \text{train-classifier}(Z)$ 
    \State \textbf{Return} $\lambda$
\EndFunction
\end{algorithmic}
\end{algorithm}

\begin{algorithm}
\caption{Prediction}\label{alg:prediction}
\begin{algorithmic}[1]
\Function{Prediction}{$x_q$, $\mathbf{e}$, $f$, $\lambda$}
    \For{view $v$ = 1 to $n$}
        \State $x_v = \text{feature-extraction}_v(x_q)$
        \State $z_v = e_v(x_v)$ 
    \EndFor
    
    \State $z = f(z_1, z_2, \dots, z_n)$ 
    
    \State $\hat{y} = \lambda(z)$
    
    \State \textbf{Return} $\hat{y}$
\EndFunction
\end{algorithmic}
\end{algorithm}

\section{Experimental Methodology}
\subsection{Datasets}

Table~\ref{tab:datasets} shows three fake news datasets (FAKES, LIAR, and ISOT) used in the experiments. All datasets have news in English and formal language and contain labels to identify the fake news. The LIAR dataset is evaluated both as a binary classification and as a six-class classification task.

\begin{table}[H]
\centering
\begin{tabular}{llll}
\textbf{Dataset}                                & \textbf{Domain}         & \textbf{No. Classes} & \textbf{Samples} \\ \hline
FAKES \cite{salem2019fa}       & War in Syria            & 2                & 804              \\
LIAR \cite{wang2017liar}       & Politics                & 2                & 12791            \\
LIAR \cite{wang2017liar}                   & Politics                & 6                & 12791            \\
ISOT \cite{ahmed2017detection} & Politics and World news & 2                & 44896            \\ \hline
\end{tabular}
\caption{Datasets description}
\label{tab:datasets}
\end{table}




\subsection{Views generation}

The generation of individual views is crucial for the proposed model process. These views act as the input to the autoencoder, a key component that generates a new representation. The process of generating individual views is a two-step process, involving preprocessing and feature extraction.

In the preprocessing step, we standardize the text by reducing language variations and removing terms or characters that do not contribute significant information. Specifically, we remove URLs, special characters (such as punctuation marks and symbols), stopwords (commonly used words that do not carry meaningful context, like "the," "is," "at"), and words that appear only once in the corpus. Additionally, we apply lemmatization to reduce words to their base or root form (e.g., "running" becomes "run").

In the feature extraction step, the text is transformed into numeric vectors. Techniques based on frequency, such as Bag-of-Words, also known as Count Vectorizer (CV), and TF-IDF, word embedding methods like Word2Vec \cite{mikolov2013efficient}, GloVe \cite{pennington2014glove}, and FastText \cite{joulin2016fasttext}, as well as language models like RoBERTa \cite{liu2019roberta} and Falcon \cite{almazrouei2023falcon}, were utilized.
The selected textual representation methods are widely used in fake news detection tasks and have achieved strong performance.~\cite{farhangian2024fake}.


\subsection{Experiments setup}

In this paper, we are utilizing the following as the basis for evaluation:
\begin{itemize}
    \item Views: TF-IDF, CV, Word2Vec, Glove, Fast, RoBERTa, and Falcon.
    \item Multi-view autoencoder models: jointAAE, wAAE, DVCCA, me\_mVAE, MoPoEVAE, and mvtCAE.
    \item Latent dimension sizes: 7, 21, 70, 350, 700, and 3500.
    \item Classifiers: Logistic regression, SVM, Random Forest, Naïve Bayes, MLP, Extremely randomized trees (Extra), and KNN.
    \item Datasets: FAKES, LIAR (2 and 6 classes), and ISOT.
\end{itemize}

The views (feature representations) and classifiers were selected based on the analyses conducted in Farhangian et al.~\cite{farhangian2024fake} work, which demonstrates that the selected models are widely used in the literature and exhibit strong performance in the task of fake news detection. The multi-view autoencoder models were chosen based on Aguila et al.~\cite{aguila2023multi} work, which presents and provides several multi-view autoencoders. However, only models that included a joint representation layer were selected, as our objective is to use this joint representation as input for the classifier.

We conducted two main sets of experiments:
\begin{itemize}
    \item Execution of all multi-view autoencoder models, varying the latent vector dimension size with each classifier, using the entire set of views as input.
    \item Execution of a single multi-view autoencoder model using all possible combinations of views as input.
\end{itemize}

All experiments are run with all databases. Except for the LIAR dataset, which already has pre-defined training and test sets, we split the datasets into 70\% for training and 30\% for testing.




\section{Results and discussion}

Looking to answer our research questions, we evaluate (1)~the performance of different multi-view autoencoder models in the representation and detection of fake news, (2)~the effectiveness of multi-view approaches compared to individual views, and (3)~the relevance of each view in contributing to the classification task.


Firstly, we evaluate the multi-view autoencoder models (jointAAE, wAAE, DVCCA, me\_mVAE, MoPoEVAE, and mvtCAE) by varying the latent dimension size (7, 21, 70, 350, 700, and 3500) and using different classification models (Logistic regression, SVM, Random Forest, Naïve Bayes, MLP, Extra, and KNN). In this experiment, we use all views (TF-IDF, CV, W2V, Glove, Fast, and Falcon) as input to the multi-view autoencoders. 

\begin{figure}[]
\centering
\begin{subfigure}[t]{0.24\textwidth}
  \centering
  \includegraphics[width=\linewidth]{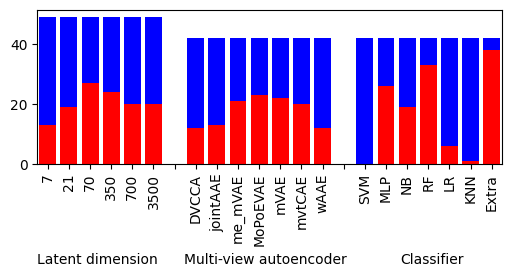}
  \caption{FAKES}
  \label{fig:FAKES_count_exe}
\end{subfigure}
\begin{subfigure}[t]{0.24\textwidth}
  \centering
  \includegraphics[width=\linewidth]{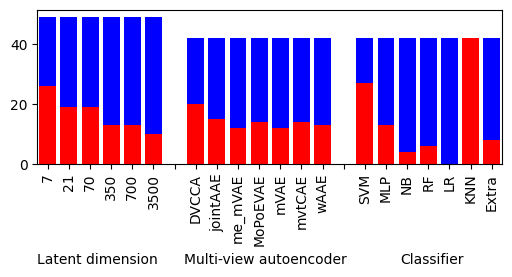}
  \caption{LIAR 2 classes}
  \label{fig:LIAR_2_count_exe}
\end{subfigure}
\begin{subfigure}[t]{0.24\textwidth}
  \centering
  \includegraphics[width=\linewidth]{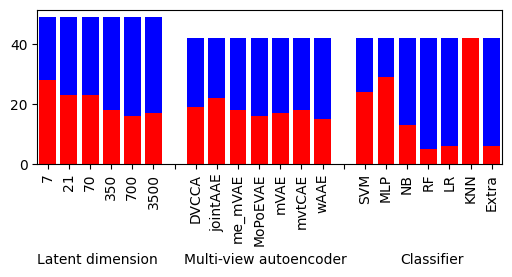}
  \caption{LIAR 6 classes}
  \label{fig:LIAR_6_count_exe}
\end{subfigure}
\begin{subfigure}[t]{0.24\textwidth}
  \centering
  \includegraphics[width=\linewidth]{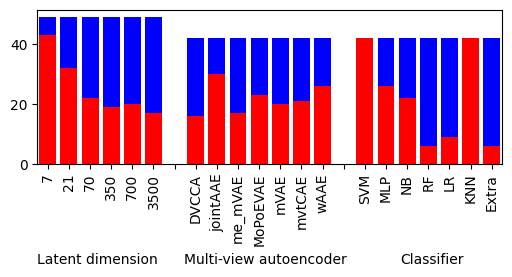}
  \caption{ISOT}
  \label{fig:ISOT_count_exe}
\end{subfigure}
\caption{Number of configurations below (red) and equal to or above (blue) the
average of all possible configurations.}
\label{fig:count_exe}
\end{figure}

Since we have six latent dimension sizes, seven multi-view autoencoder models, and seven classifiers, we obtain a total of 294 ($6 \times 7 \times 7$) configurations.
Figure~\ref{fig:count_exe} shows the number of configurations that obtained accuracy below (red) and equal to or above (blue) the average of all 294 configurations. This figure analyzes three different components for each dataset: ``Latent dimension'', ``multi-view autoencoder'', and ``Classifier''. For instance, the ``Latent dimension'' component has 6 bars, each with a size of 49 ($7 \text{ multi-view autoencoder models} \times 7 \text{ classifiers}$). When analyzing the ``Latent dimension'', a noticeable trend suggests that larger latent dimension sizes tend to perform above the average more frequently. This indicates that smaller latent dimensions tended to capture less relevant information, resulting in poorer performance. However, excessively large dimensions can not necessarily yield further gains and sometimes cause overfitting. There is, however, a middle ground where the latent dimensions provide sufficient representation capacity while maintaining generalization across the classifiers.

\begin{figure}[]
\begin{subfigure}[t]{0.24\textwidth}
  \centering
  \includegraphics[width=\linewidth]{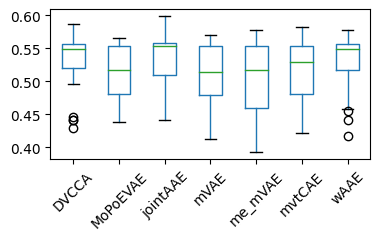}
  \caption{FAKES}
  \label{fig:FAKES_boxplot_ae}
\end{subfigure}
\begin{subfigure}[t]{0.24\textwidth}
  \centering
  \includegraphics[width=\linewidth]{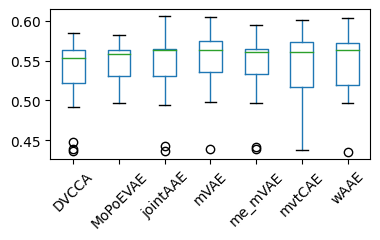}
  \caption{LIAR - 2 classes}
  \label{fig:LIAR_2_boxplot_ae}
\end{subfigure}
\begin{subfigure}[t]{0.24\textwidth}
  \centering
  \includegraphics[width=\linewidth]{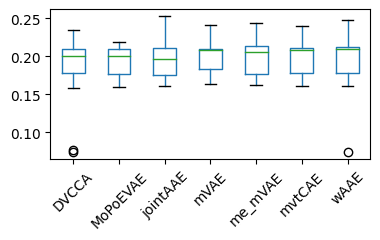}
  \caption{LIAR - 6 classes}
  \label{fig:LIAR_6_boxplot_ae}
\end{subfigure}
\begin{subfigure}[t]{0.24\textwidth}
  \centering
  \includegraphics[width=\linewidth]{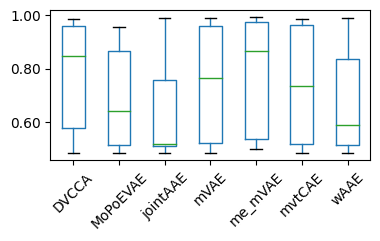}
  \caption{ISOT}
  \label{fig:ISOT_boxplot_ae}
\end{subfigure}
\caption{Boxplot of the accuracy of the executions of latent sizes and classifiers grouped by multi-view autoencoder model}
\label{fig:boxplot_ae}
\end{figure}

Still analyzing Figure \ref{fig:count_exe}, it is clear that execution performance is highly sensitive to the type of classifier used. Some classifiers, regardless of the multi-view model employed, fail to learn the task. When examining the multi-view autoencoder models, it is noted that they are statistically equivalent. This is more apparent when observing the boxplot of executions for each multi-view autoencoder model presented in Figure \ref{fig:boxplot_ae}. Since they are distinct models, it is expected that they would learn different representations. However, in general, all the multi-view autoencoder models successfully created a joint representation that encapsulates information from the views, thus answering the research question (1). Nevertheless, performance is superior depending on the choice of latent dimension size and classifier.

\begin{figure}[]
\begin{subfigure}[t]{0.49\textwidth}
  \centering
  \includegraphics[width=\linewidth]{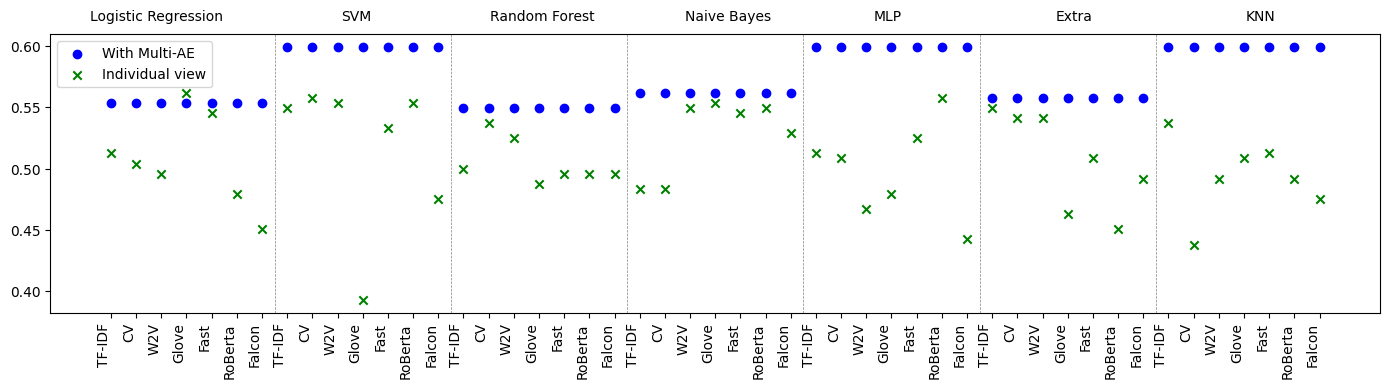}
  \caption{FAKES}
  \label{fig:FAKES_ind_X_multiae}
\end{subfigure}
\begin{subfigure}[t]{0.49\textwidth}
  \centering
  \includegraphics[width=\linewidth]{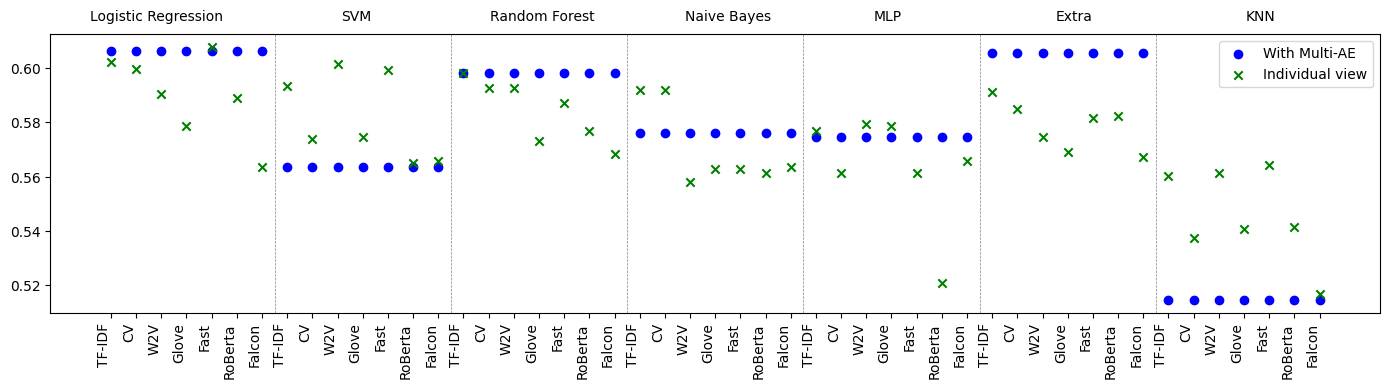}
  \caption{LIAR 2 classes}
  \label{fig:LIAR_2_ind_X_multiae}
\end{subfigure}
\begin{subfigure}[t]{0.49\textwidth}
  \centering
  \includegraphics[width=\linewidth]{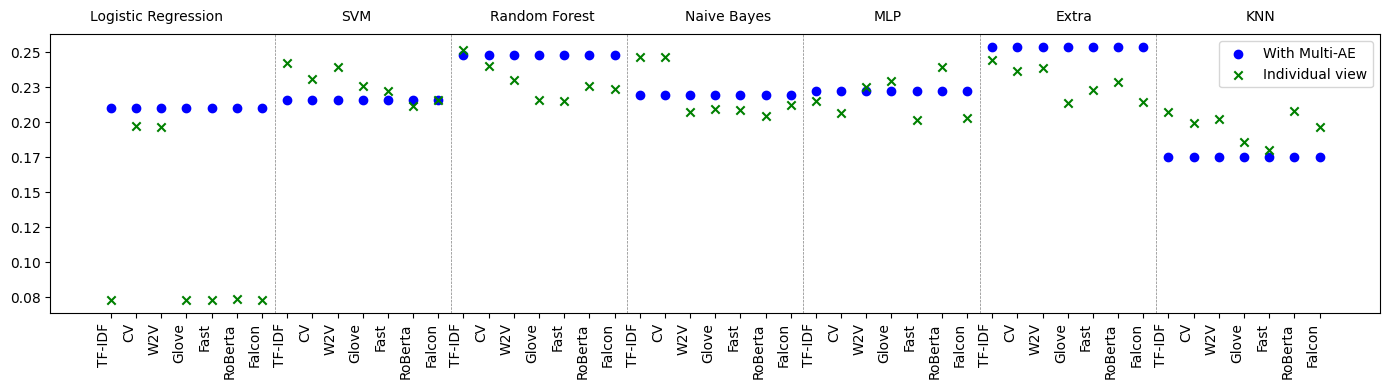}
  \caption{LIAR 6 classes}
  \label{fig:LIAR_6_ind_X_multiae}
\end{subfigure}
\begin{subfigure}[t]{0.49\textwidth}
  \centering
  \includegraphics[width=\linewidth]{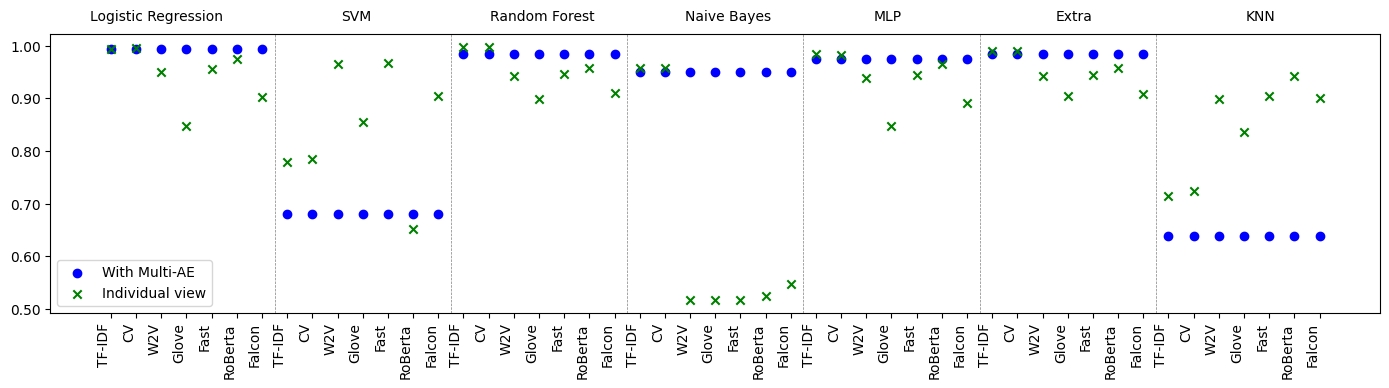}
  \caption{ISOT}
  \label{fig:ISOT_ind_X_multiae}
\end{subfigure}
\caption{Comparison of each view with each classifier versus the multi-view autoencoder model that obtained the best result with the same classifier and contains the view in its input set}
\label{fig:all_ind_X_multiae}
\end{figure}

To compare the proposed approach performance, we built a baseline with the classifiers and individual's views/feature extractions. 
When we individually compare each view with the multi-view models containing the same view within the input set, we can observe that the multi-view approach achieves superior performance in almost all classifiers.
Figure~\ref{fig:all_ind_X_multiae} presents the comparison of the accuracy of each individual feature extraction combined with each classifier (green x) and compares it with the best multi-view autoencoder model that included the specific view in its input set (blue dot). This visualization clearly demonstrates that in nearly all experiments, the individual view consistently underperforms compared to the multi-view approach, and this answers research question (2).

\begin{figure*}[]
\begin{subfigure}[t]{0.99\textwidth}
  \centering
  \includegraphics[width=\linewidth]{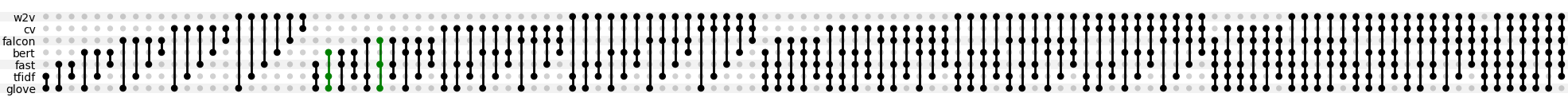}
  \caption{FAKES}
  \label{fig:FAKES_comb_all_views}
\end{subfigure}
\begin{subfigure}[t]{0.99\textwidth}
  \centering
  \includegraphics[width=\linewidth]{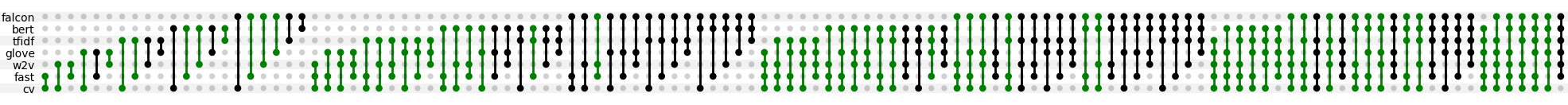}
  \caption{LIAR 2 classes}
  \label{fig:LIAR_2_comb_all_views}
\end{subfigure}
\begin{subfigure}[t]{0.99\textwidth}
  \centering
  \includegraphics[width=\linewidth]{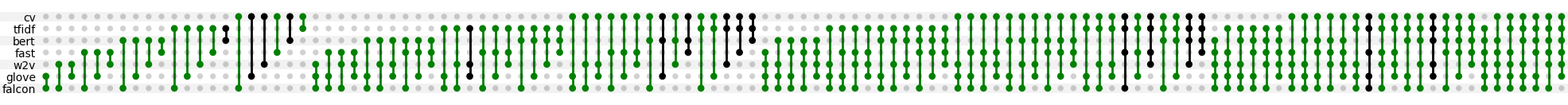}
  \caption{LIAR 6 classes}
  \label{fig:LIAR_6_comb_all_views}
\end{subfigure}
\begin{subfigure}[t]{0.99\textwidth}
  \centering
  \includegraphics[width=\linewidth]{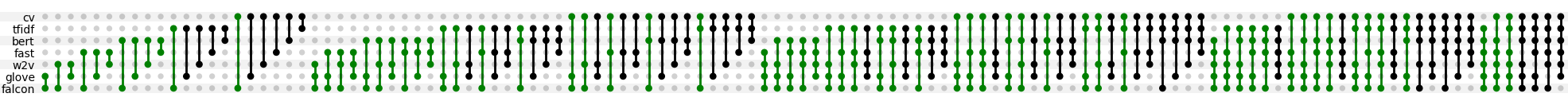}
  \caption{ISOT}
  \label{fig:ISOT_comb_all_views}
\end{subfigure}
\caption{F-score comparison between using all possible combination views and all views using jointAAE multi-view autoencoder with dimension latent 70, where the green columns represent that the combination of views achieved a better result than using all the views.}
\label{fig:comb_all_views}
\end{figure*}

The results presented have utilized all views as input. So, we question the necessity of including all views to obtain a comprehensive representation of the news. Given that, we conducted experiments using all possible combinations of input sets (views) in the multi-view autoencoder model. For this experiment, we opted for the jointAAE model with a latent dimension of 70, which was the model that obtained the best result with the FAKES dataset. Figure \ref{fig:comb_all_views} illustrates the combination of all views. Each row represents a view, and each column corresponds to an experiment, with the views used as input marked by a dot. In this figure, the experiments that achieved better results than the ones that used all views as input are highlighted in green. For all datasets, at least one subset of views outperformed the experiment that used all views as input. In fact, the majority of the experiments achieved better results with a subset of views than with all views, except for the FAKES dataset, where only two views were better than using all views. This answers research question (3), indicating that using a more selective combination of views leads to better performance than relying on all views, suggesting that not all features contribute equally to the model's effectiveness.


\section{Conclusion}

This work assembled different views into one single representation using multi-view autoencoders for fake news detection. This new view has the advantage of maintaining the most relevant information about each original view into a single representation. 


The experiments evaluated different fake news datasets, feature representation algorithms, latent dimensions, classifiers, and multi-view autoencoders. We observed that multi-view approaches consistently overcome individual views. No overall guidance highlights an optimal choice regarding the latent dimension and classifier. However, the jointAAE model with a latent dimension of 70 can be highlighted. 

Using all available views to construct the new representation eliminates the need to choose which view best fits the task. Although using all views as input to the multi-view autoencoder yields good performance, using a subset of these views performs even better.


For future work, we plan to develop an automatic view selection mechanism and evaluate this approach in other domains, such as hate speech detection and sentiment analysis.



\subsection{Limitations}


One of the primary limitations of the current study is the focus on a single modality, specifically text, in the multi-view setting. A potential improvement lies in extending the approach to incorporate multiple modalities. The multi-view autoencoder framework, being modality-agnostic, could be employed to integrate diverse sources of information such as images, networks, and text. Using multiple modalities could enhance the model's ability to capture more complex and comprehensive patterns in the data. One approach to the multi-view autoencoder is that the latent layer condenses all views into a single data vector.

Another limitation is that, despite the existence of several text feature extraction methods (such as word embeddings and language models), we evaluated only a limited set of these methods, selecting those we considered most relevant to initiate the study. We plan to conduct evaluations using more views as input for future work.






\end{document}